\documentclass[sigconf]{acmart}
\AtBeginDocument{%
  }

\setcopyright{acmlicensed}
\copyrightyear{2018}
\acmYear{2018}
\acmDOI{XXXXXXX.XXXXXXX}

\acmConference[Conference acronym 'XX]{Make sure to enter the correct
  conference title from your rights confirmation email}{June 03--05,
  2018}{Woodstock, NY}
\acmISBN{978-1-4503-XXXX-X/2018/06}
\usepackage{graphicx}
\usepackage[table]{xcolor}
\usepackage{multirow}
\usepackage{makecell}

\begin{document}

\title{RedOne 2.0: Rethinking Domain-specific LLM Post-Training in Social Networking Services}

\author{Fei Zhao, Chonggang Lu, Haofu Qian, Fangcheng Shi, Zijie Meng, Jianzhao Huang,\\Xu Tang, Zheyong Xie, Zheyu Ye, Zhe Xu, Yao Hu, Shaosheng Cao}
\authornote{Corresponding author.}
\affiliation{%
  \institution{NLP Team, Xiaohongshu Inc.}
  \city{Huangpu District}
  \state{Shanghai}
  \country{China}
}
\email{caoshaosheng@xiaohongshu.com}

\renewcommand{\shortauthors}{Fei Zhao et al.}

\begin{abstract}
    As a key medium for human interaction and information exchange, social networking services (SNS) pose unique challenges for large language models (LLMs): heterogeneous workloads, fast-shifting norms and slang, and multilingual, culturally diverse corpora that induce sharp distribution shift. Supervised fine-tuning (SFT) can specialize models but often triggers a ``seesaw'' between in-distribution gains and out-of-distribution robustness, especially for smaller models. To address these challenges, we introduce RedOne 2.0, an SNS-oriented LLM trained with a progressive, RL-prioritized post-training paradigm designed for rapid and stable adaptation. The pipeline consist in three stages: (1) Exploratory Learning on curated SNS corpora to establish initial alignment and identify systematic weaknesses; (2) Targeted Fine-Tuning that selectively applies SFT to the diagnosed gaps while mixing a small fraction of general data to mitigate forgetting; and (3) Refinement Learning that re-applies RL with SNS-centric signals to consolidate improvements and harmonize trade-offs across tasks. Across various tasks spanning three categories, our 4B scale model delivers an average improvements about 2.41 over the 7B sub-optimal baseline. Additionally, RedOne 2.0 achieves average performance lift about  8.74 from the base model with less than half the data required by SFT-centric method RedOne, evidencing superior data efficiency and stability at compact scales. Overall, RedOne 2.0 establishes a competitive, cost-effective baseline for domain-specific LLMs in SNS scenario, advancing capability without sacrificing robustness.
\end{abstract}



\received{20 February 2007}
\received[revised]{12 March 2009}
\received[accepted]{5 June 2009}

\maketitle

\section{Introduction}
Powered by the rapid advance of electronic devices and network infrastructure, social networking services (SNS) have evolved into core infrastructure for human daily interaction and the spread of information~\cite{xia2013socially}. In parallel, large language models (LLMs) have delivered striking progress in natural language processing (NLP), enabling promising performance across a wide range of downstream tasks~\cite{gpt4o, yang2025qwen3}. However, deploying a general-purpose LLM into an SNS scenario is not a simple lift-and-shift. The workloads of SNS platform are highly heterogeneous, such as real-time moderation and abuse response, recommendation-shaped dialogue, creator assistance, and community operations, each with distinct latency, safety, and tone requirements. And the environment changes in a very fast speed, where trends, slang, and community norms rise and fade within days. Additionally, because SNS connects vast, diverse audiences, models must handle large, complex corpora across languages and expression styles to bridge users from different cultural backgrounds. Together these factors amplify distribution shift and increase the risk of brittle generalization, where models, optimized on the standard benchmarks, may misread community-specific rules, over- or under-enforce policies, or drift as conventions change~\cite{zhao2025redone,zeng2024llmforsocialnetworks}. Therefore, seamlessly integrating LLMs for SNS scenario requires that adapt quickly, remain stable, and maintain competence across languages and communities, especially without compromising safety or user trust.

\begin{figure}[t]
    \centering
    \includegraphics[width=\linewidth]{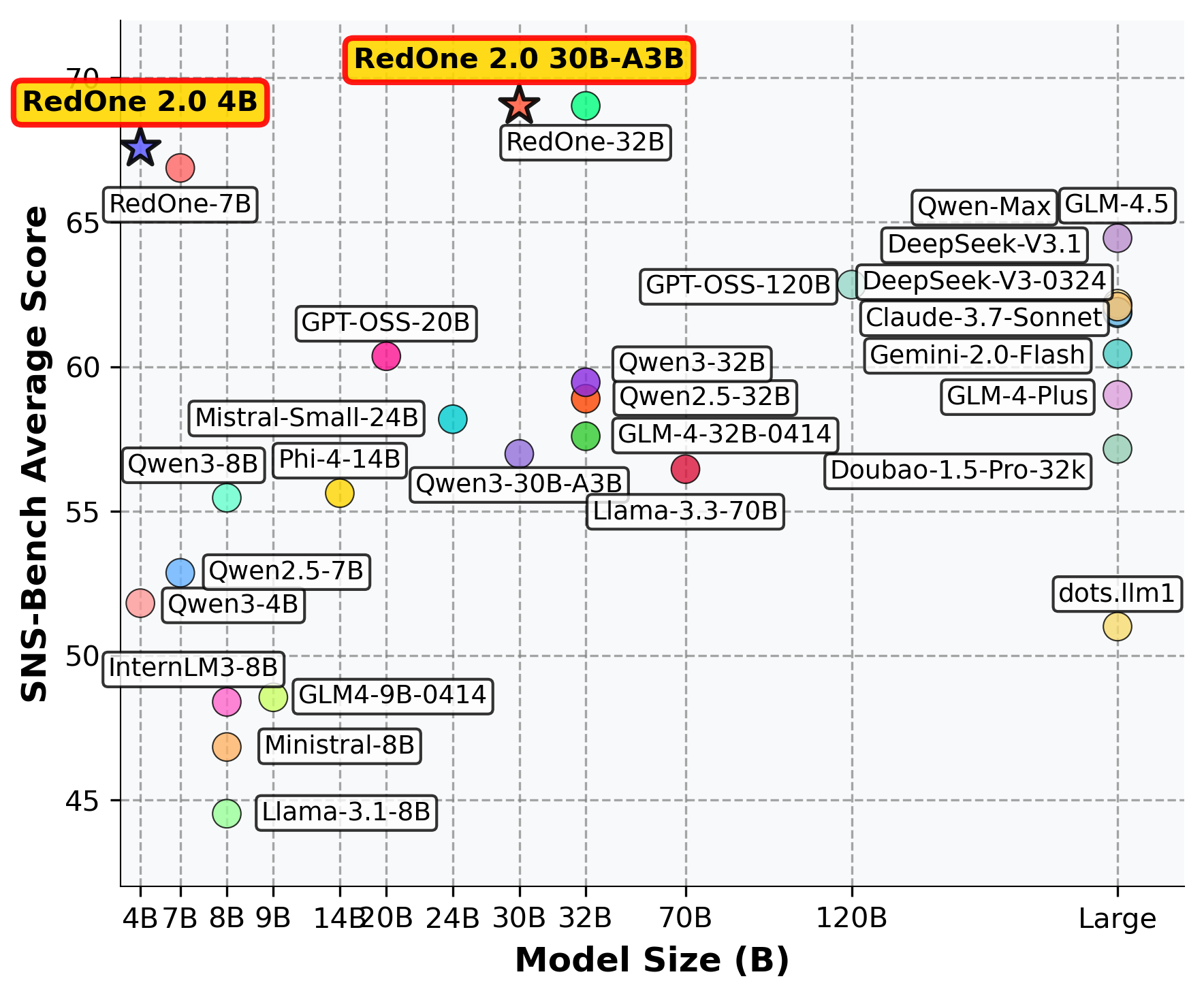}
    \caption{The comparison of various scale models' performance in the SNS domain.}
    \label{fig:cmp}
\end{figure}

As a pioneer in adapting LLM in SNS, RedOne~\cite{zhao2025redone} established an early domain-specific post-training route applying a classical SFT-based pipeline and showed that SNS performance can be increased without sacrificed general ability. However, this SFT-dominated pipelines exhibit a pronounced ``seesaw'' effect, which improves on in-distribution (ID) tasks often come at the expense of out-of-distribution (OOD) generalization. And the problem is especially acute for smaller models, which are more susceptible to catastrophic forgetting as new domain patterns overwrite previously learned skills. General mitigations try to smooth the issue by scaling huger training sets, crafting elaborate data mixture schedules, and designing diverse learning objectives~\cite{kotha2023understanding,kumar2022fine,ramasesh2021effect,huang2024mitigating}. While these methods can somehow recover models' robustness on OOD dataset, they lift up the budgets of data and compute to a sky-high level. In contrast, reinforcement learning (RL) offers a more distinctive advantage for domain specific post-training~\cite{jin2025rl}. By optimizing against preference or reward signals, RL directly aligns model behavior with human and downstream objectives, preserving existing competencies while unlocking latent capabilities. As RL has matured in LLM optimization, its practical value has been demonstrated across preference alignment, safety shaping, controllable generation and task-level policy tuning~\cite{ouyang2022training,deepseekr1,yu2025dapo}. However, how to structure RL-centric training to balance data efficiency, stability, and domain transfer in the fast-evolving SNS environment remains underexplored.

To address this gap, we introduces RedOne 2.0, a new SNS-oriented LLM that adopts a progressive, RL-prioritized post-training paradigm. Our design philosophy is that exploration targeted correction refinement yields better stability generalization trade-offs than SFT-heavy recipes, especially at smaller parameter scales and with limited domain data, as shown in Fig.~\ref{fig:cmp}. Concretely, RedOne 2.0 is trained in three stages: 1) Exploratory Learning. The model is exposed to curated SNS corpora to establish initial domain alignment and to diagnose the lack of ability for realistic distributions. 2) Targeted Fine-Tuning. We apply SFT on tasks where previous stage diagnostics reveal systematic weaknesses, blending a small fraction of general data to explicitly regularize against forgetting and retain broad generalization. 3) Refinement Learning. Building on the corrected model, we reuse RL with SNS-centric signals to consolidate improvements and smooth trade-offs across different tasks, yielding the coherent capability gains.

We also conducted extensive experiments across various tasks spanning three categories to validate our post-training pipeline. The 4B-parameter variant surpasses the 7B counterpart by 2.41 on average, demonstrating that strong performance is attainable at compact scales. Using Qwen3-4B as the base, RedOne 2.0 requires only half of RedOne's data while achieving a performance lift with 8.74, which demonstrates superior data efficiency and broader capability gains from RL-centric curriculum learning. 

Our contribution can be summarized as follows:

\begin{itemize}
    \item We present RedOne 2.0, an SNS-domain LLM that achieves higher capability with less data and smaller models.
    \item We propose a progressive, RL-prioritized post-training paradigm that delivers consistent improvements in both general and SNS-specific abilities while mitigating the ``seesaw'' effect of SFT.
    \item We provide comprehensive empirical validation showing state-of-the-art results in the SNS scenario and strong robustness under distribution shift, establishing RedOne 2.0 as a competitive and cost-effective baseline for domain LLMs.
\end{itemize}

\begin{figure*}[t]
    \centering
    \includegraphics[width=\textwidth]{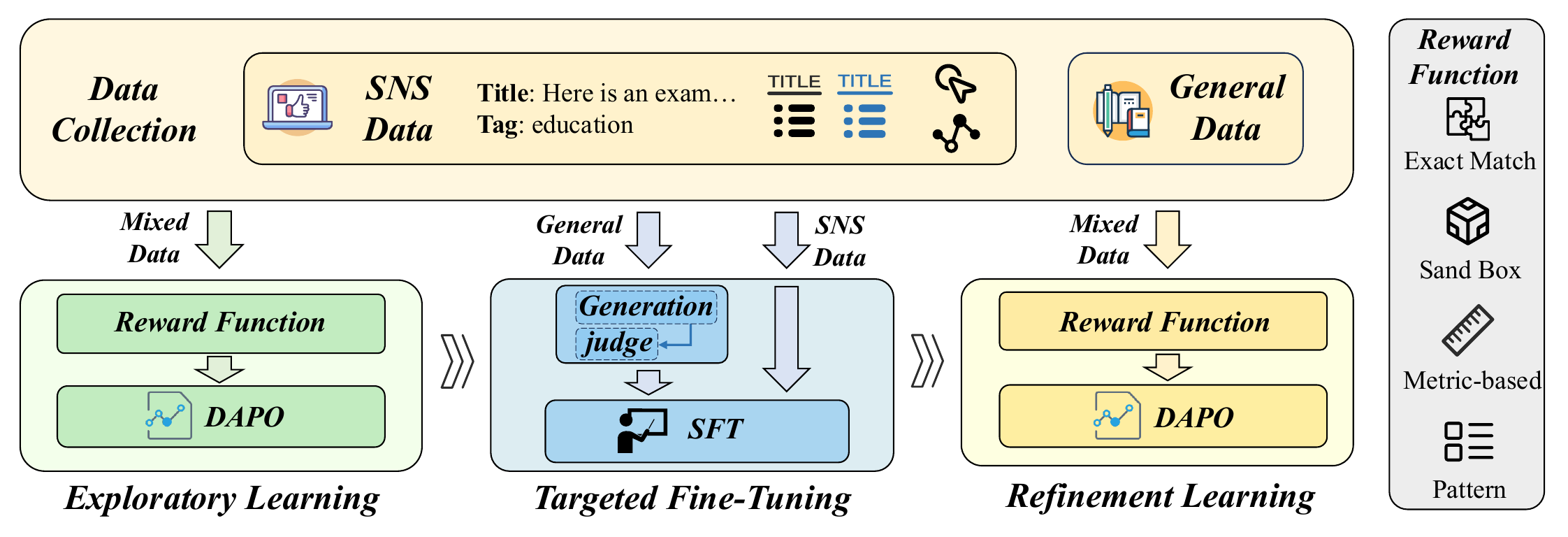}
    \caption{Overview of our RL-based incremental three-stage training pipeline.}
    \label{fig:main}
\end{figure*}

\section{Related Work}
\subsection{LLM in Social Networking Services}
Given the central role of SNS platforms in everyday information exchange, interest in this domain has surged~\cite{bakshy2015exposure,vosoughi2018spread,altay2025following}. Advances in LLMs have accelerated integration across sentiment analysis~\cite{zhang2023sentiment,wang2023reducing}, search and question answering over social content~\cite{shah2024stackeval,kahl2024llms}, personalized content generation~\cite{lubos2024llm,lyu2023llm}, content moderation~\cite{kumar2024watch,kolla2024llm}, and platform operations~\cite{feng2024does,qiao2025botsim}. Moving beyond task-specific pipelines that underuse model knowledge and generalization, recent work explores unified solutions. Social-LLM combines localized social interaction signals with text features to scale inductive user detection across seven real-world datasets~\cite{jiang2023socialllm}. \cite{zeng2024llmforsocialnetworks} organizes LLM-for-SNS applications into knowledge, engagement, and foundation tasks while outlining deployment challenges. RedOne introduces a large-scale SNS dataset and a common domain post-training recipe that yields strong offline gains and measurable online improvements~\cite{zhao2025redone}. Yet most approaches remain data-centric, expanding domain competence primarily through large annotated corpora and dependence on strong base models, which drives up cost. RedOne 2.0 revisits post-training in the SNS setting and achieves stronger downstream performance with substantially less data and smaller model scales.

\subsection{General LLM Post-training}
Post-training bridges large-scale pre-training and deployment by enhancing instruction following, safety, and factuality. Typical pipelines first perform SFT on curated instruction–response pairs, then apply preference optimization with human or automated feedback, as exemplified by InstructGPT~\cite{ouyang2022training}. Subsequent advances such as RRHF~\cite{yuan2023rrhf} and DPO~\cite{rafailov2023direct} simplify preference learning and improve training stability, while GRPO~\cite{shao2024deepseekmath} and DAPO~\cite{yu2025dapo} introduce more efficient, reward-driven reinforcement learning frameworks that better balance exploration and alignment quality. Despite these developments, most approaches remain domain-agnostic and tend to underutilize specialized knowledge crucial for vertical applications.

\subsection{Domain-specific LLM Post-training}
Domain adaptation injects targeted knowledge and preferences to boost in-domain performance. Typical pipelines combine continued pre-training on domain corpora with supervised fine-tuning and preference optimization on domain tasks, yielding strong results in finance~\cite{wu2023bloomberggpt,bhatia2024fintral,yang2023investlm,gruca2008incentive}, law~\cite{guha2023legalbench,katz2024gpt,dahl2024large,colombo2024saullm}, medicine~\cite{singhal2025toward,xu2025lingshu,liu2025dentvlm,jiang2025omniv}, and the sciences~\cite{zhang2024sciglm,azerbayev2023llemma,taylor2022galactica,bai2025intern}. However, heavy reliance on SFT can overfit to benchmarks and weaken generalization, while later reinforcement learning often only partly fixes this drift. RedOne 2.0 addresses these limits with an RL-centric design that emphasizes staged reinforcement and dynamic task sampling to improve both general competence and domain reasoning.

\section{Methodology for RedOne 2.0}
As illustrated in Fig.~\ref{fig:main}, the post-training pipeline of RedOne 2.0 includes three stages, and each stage builds on the previous one, progressively aligning the model to the SNS domain while preserving general capabilities and ensuring stability. First, during Exploratory Learning, we initially align the base model to the SNS domain and identify SNS-specific tasks where the model requires targeted reinforcement. Next, in Targeted Fine-Tuning, we address the deficiencies revealed in the previous stage to improve model's performance in SNS scenario and create exploration space for subsequent optimization. Finally, in Refinement Learning, we apply RL on SNS-domain data to stabilize and smooth model behavior, driving better performance across various tasks.

\subsection{Dataset Definition}
To ground the three-stage pipeline, we first construct and characterize the training dataset. We curate large-scale data from both the SNS domain and the general domain. The former covers capabilities commonly evaluated in SNS scenarios, including information extraction, semantic matching, content understanding, user behavior modeling, dialogue, and translation, and spans more than 75 carefully defined tasks. This provides sufficient supervision for following training pipeline and establishes a solid data foundation for adapting RedOne 2.0 to a wide range of SNS real-world applications. The latter integrates high-quality open-source datasets that have been widely validated by the community, thereby reducing redundant annotation and processing costs while ensuring robust general capabilities of RedOne 2.0.

We then normalize all collected data into a unified format of question $Q$ and answer $A$, yielding the final dataset
\begin{equation}
    \mathcal{D} = \{\, \mathcal{D}_{\text{SNS}} \cup \mathcal{D}_{\text{GEN}} \mid (Q, A) \,\}
\end{equation}
where $\mathcal{D}_{\text{SNS}}$ and $\mathcal{D}_{\text{GEN}}$ denote the SNS-domain and general-domain subsets, respectively.

\subsection{Exploratory Learning}
The goal of Exploratory Learning is to achieve initial alignment between the base model and SNS-domain characteristics. Instead of committing early to narrowly scoped objectives, this stage immerses the model in diverse SNS data to capture the breadth of task distributions and domain-specific interaction patterns, while preserving general competence and revealing tasks that remain hard due to unfamiliarity from base model.

\subsubsection{Data Construction}
We select about 750K SNS entries $\mathcal{D}_{\text{SNS}_1}$ from $\mathcal{D}_{\text{SNS}}$, covering 75 heterogeneous tasks and all capability types, such as post taxonomy, query classification, machine reading comprehension, post view search and SNS domain translation. A balanced sampling schedule is also conducted to preserve the visibility of long-tail behaviors. Additionally, to maintain reasoning and general competence, we attach 50K data $\mathcal{D}_{\text{GEN}_1}$ from $\mathcal{D}_{\text{GEN}}$ with rationales, which are widely regarded as beneficial for preserving model knowledge and supporting structured reasoning.

\subsubsection{Reward Function}
In RL, the reward function is the most critical supervision signal during training. Unlike prior works~\cite{deepseekr1} that takes a simple approach by defining a single rule-based reward to check whether a response follows a specified format and whether the final answer is correct, we take a different path. Considering that the downstream scenarios aligned with RedOne 2.0 span heterogeneous tasks with substantial variation in both format and content, and that their evaluation criteria are highly diverse, we define task type specific rewards for a sampled pair $(Q, A)$ and model's output $O$ as following:

1) \textbf{Exact Match.} For close-ended problems with determinate answers, such as classification or multiple-choice, we focus on constraining answer consistency with exact match score.
\begin{equation}
    \mathcal{R}_{\text{EM}}(O,A)=
    \begin{cases}
    1, & O = A, \\
    0, & \text{otherwise}.
    \end{cases}
\end{equation}
    
2) \textbf{Metrics-based.} For open-ended tasks such as translation, we avoid seeking a single binary ``correct'' standard. Instead, we define rewards using task-specific evaluation metrics $\text{Eval}$.
\begin{equation}
    \mathcal{R}_{\text{Met}}(O,A)=\text{Eval}(O,A)
\end{equation}

3) \textbf{Sand Box.} For tasks like code generation, traditional exact-match or metric-based scoring struggles to objectively assess output quality. The most direct approach is sandbox simulation, where we create an execution environment to run the generated solution and evaluate it by the obtained results $\text{Exe}(O)$.
\begin{equation}
    \mathcal{R}_{\text{SandBox}}(O,A)=
    \begin{cases}
    1, & \text{Exe}(O) = A, \\
    0, & \text{otherwise}.
    \end{cases}
\end{equation}

4) \textbf{Pattern.} Additionally, given the instability of generative LLM output formats and the existence of community benchmarks targeting instruction following, we design a pattern-based matching mechanism $\text{Match}$ that emphasizes adherence to specified formats rather than the semantic content itself.
\begin{equation}
    \mathcal{R}_{\text{Pattern}}(O,A)=\text{Match}(O,A)
\end{equation}

Finally, each sample's reward $\mathcal{R}(O,A)$ is mapped to the corresponding reward function based on its task category.

\subsubsection{Domain Alignment}
In this stage, our domain alignment objective is to jointly raise general competence and SNS-specific capability while fully exploiting the model's latent potential, and to systematically surface sub-tasks that remain difficult due to base model's unfamiliarity. In practice, we randomly mix $\mathcal{D}_{\text{SNS}_1}$ and $\mathcal{D}_{\text{GEN}_1}$ to form $\mathcal{D}_1$, which conducts DAPO-based~\cite{yu2025dapo} RL training for this stage. For a specific data entry $(Q, A)$, DAPO samples a group of $G$ individual candidate outputs $\{O_i\}_{i=1}^G$ from the old policy model $\pi_{\theta_{\mathrm{old}}}$. Then, we can optimize the policy $\pi_{\theta}$ by optimizing the following loss function:
\begin{align}
    \mathcal{L}_{\mathrm{DAPO}}(\theta)
    =&\mathbb{E}_{(Q,A)\sim \mathcal{D}_1, \{O_i\}_{i=1}^G \sim \pi_{\theta_{\mathrm{old}}}(\cdot \mid Q)} \nonumber\\
    &\Biggl[
    \frac{1}{\sum_{i=1}^G |O_i|}
    \sum_{i=1}^G \sum_{t=1}^{|O_i|}
    \min\Bigl(r_{i,t}(\theta)\,\hat A_{i,t}, \nonumber\\
    & \mathrm{clip}(r_{i,t}(\theta),\,1-\varepsilon_{\mathrm{low}},\,1+\varepsilon_{\mathrm{high}})\,\hat A_{i,t}\Bigr)
    \Biggr] \nonumber\\
    &\text{s.t.}\quad 0 < \bigl\lvert\{O_i \mid \mathrm{is\_equivalent}(A, O_i)\}\bigr\rvert < G
\end{align}
where $\varepsilon_{\mathrm{low}}$ and $\varepsilon_{\mathrm{high}}$ control the clipping range, and
\begin{equation}
    r_{i,t}(\theta) = \frac{\pi_\theta(O_{i,t} \mid Q, O_{i,<t})}{\pi_{\theta_{\mathrm{old}}}(O_{i,t} \mid Q, O_{i,<t})}
\end{equation}

\begin{equation}
    \hat A_{i,t} = \frac{\mathcal{R}_i - \mathrm{mean}(\{\mathcal{R}_i\}_{i=1}^G)}{\mathrm{std}(\{\mathcal{R}_i\}_{i=1}^G)}
\end{equation}

Finally, this alignment stage yields broad, stable gains without premature specialization, while producing a fine-grained diagnosis of where the model underperforms. And the resulting capability map guides targeted repair in the next stage.

\subsection{Targeted Fine-Tuning}
After initial SNS alignment, Targeted Fine-Tuning directly addresses tasks that remain weak. The emphasis is on repairing deficiencies while preserving previous gains, achieved by blending difficult SNS data with carefully filtered general data.

\subsubsection{Data Preparation}
We construct a dataset $\mathcal{D}_2$ of 1.8M examples comprising 1.7M SNS instances and 100K general-domain instances. The SNS portion is derived from our pre-training data corpus $\mathcal{D}_{\text{SNS}}$, which refer to the failure tasks bucket in previous stage identified via evaluation result on various benchmarks. We further stratify these examples by capability and upweight rare but impactful cases. For the general portion, we sample them from $\mathcal{D}_{\text{SNS}}$ and introduce examples with soft labels by generating candidates responses from the previous stage model. Concretely, for a given prompt we generate 8 candidate completions by the resulted model from the first stage, and score them with a composite quality signal from a judge model, and select the best one to form a soft supervisory target. These soft labels not only mitigate catastrophic forgetting of general knowledge during SFT, but also reduce impact on distributional transformation between ``ground-truth'' labels and the first stage model's learned distribution, thereby improving learning efficiency for SNS tasks~\cite{li2017learning}. In fact, this small set of soft-labeled general data functions as a data-level regularizer, preventing SNS-focused SFT from drifting too far from the reference model.

\subsubsection{Targeted Learning}
In this stage, optimization aims to close gaps on underperforming SNS tasks while preserving gains from the previous stage. We use a plain SFT objective on a mixture of hard SNS examples and a small set of general-domain examples with soft labels:
\begin{equation}
\mathcal{L}_{\text{SFT}}=
-\,\mathbb{E}_{(Q,A)\sim \mathcal{D}_2, O \sim \pi_{\theta}(\cdot \mid Q)}
\sum_{t=1}^{|O|}\log \pi_{\theta}\!\big(O_t \,\big|\, Q, O_{<t}\big)
\end{equation}
where $\pi_{\theta}$ denotes the current policy model, $Q$ is the question, and $O$ is the target output sequence. Finally, this stage yields consistent improvements on previously weak SNS tasks while maintaining gains from the first stage on the most capability types. The combination of SNS-prioritized repair and soft-label regularization from a small general set produces a more balanced and robust model prepared for RL-based refinement in the last stage.

\begin{table*}[h]
    \centering
    \caption{Comparison results across General-Bench, SNS-Bench and SNS-TransBench. \textbf{Bold} entries indicate the best performance, while \underline{underlined} entries denote the second one in each category.}
    \label{tab:main_table}
    \resizebox{1\textwidth}{!}{%
    \begin{tabular}{l|c|cccccccc|c|cccc|c}
        \toprule
        \multirow{4}{*}{\textbf{Models}} & \textbf{General-Bench} & \multicolumn{9}{c|}{\textbf{SNS-Bench}} & \multicolumn{5}{c}{\textbf{SNS-TransBench}} \\
        \cmidrule(lr){2-2}
        \cmidrule(lr){3-11}
        \cmidrule(lr){12-16}
        {} & \multirow{2.5}{*}{\textbf{Avg.}} & \multirow{2.5}{*}{\textbf{Taxon.}} & \multirow{2.5}{*}{\textbf{Hash.}} & \multirow{2.5}{*}{\textbf{QCorr}} & \multirow{2.5}{*}{\textbf{MRC}} & \multirow{2.5}{*}{\textbf{NER}} & \multirow{2.5}{*}{\textbf{Gender}} & \multirow{2.5}{*}{\textbf{CHLW}} & \multirow{2.5}{*}{\textbf{QGen}} & \multirow{2.5}{*}{\textbf{Avg.}} & \multicolumn{2}{c}{\textbf{ZH→EN}} & \multicolumn{2}{c|}{\textbf{EN→ZH}} & \multirow{2.5}{*}{\textbf{Avg.}} \\
        \cmidrule(lr){12-15}
        & & & & & & & & & & &
        \textbf{BLEU} & \textbf{chrF++} & \textbf{BLEU} & \textbf{chrF++} & \\

        \midrule
        \multicolumn{16}{c}{\textit{Proprietary Large Language Models or The Scale of Large Language Models $>$ 100B}} \\
        \midrule
        
        GPT-4o-1120 & 70.72 & 65.79 & 84.98 & 51.79 & 58.89 & 54.99 & 88.08 & 38.96 & 47.33 & 61.35 & 40.32 & 63.91 & 49.15 & 47.28 & 50.17 \\
        Gemini-2.0-Flash & 74.42 & 68.76 & 87.36 & 48.41 & 52.21 & 53.58 & 89.64 & 37.39 & 46.27 & 60.45 & 32.72 & 58.84 & 41.80 & 40.16 & 43.38 \\
        Claude-3.7-Sonnet & 75.10 & 72.03 & 88.83 & 54.10 & 54.86 & 56.13 & 92.23 & 31.11 & 45.49 & 61.85 & 35.63 & 61.66 & 45.79 & 44.23 & 46.83 \\
        Doubao-1.5-Pro-32k & 76.13 & 30.00 & 83.21 & 58.25 & 61.32 & 56.60 & 90.67 & 30.61 & 46.55 & 57.15 & 33.71 & 61.85 & 45.54 & 44.35 & 46.36 \\
        Qwen-Max & 71.86 & 65.68 & 84.47 & 54.36 & 61.34 & 55.78 & 91.19 & 37.97 & 46.64 & 62.18 & 35.55 & 60.92 & 46.08 & 44.14 & 46.67 \\
        GLM-4-Plus & 70.25 & 65.46 & 84.31 & 52.13 & 55.81 & 53.16 & 86.53 & 30.09 & 44.68 & 59.02 & 41.57 & 65.95 & 48.79 & 47.06 & 50.84 \\
        GPT-OSS-120B & 76.71 & 67.20 & 86.04 & 56.83 & 61.45 & 55.84 & 91.19 & 38.53 & 45.61 & 62.84 & 33.06 & 59.73 & 42.67 & 40.47 & 43.98 \\
        dots.llm1 & 70.20 & 62.96 & 82.45 & 42.10 & 40.75 & 14.93 & 89.12 & 31.09 & 44.63 & 51.00 & 30.93 & 58.66 & 44.42 & 42.8 & 44.20 \\
        GLM-4.5 & 73.66 & 70.76 & 86.93 & 56.22 & 64.94 & 57.23 & 92.75 & 41.32 & 45.47 & 64.45 & 30.57 & 56.77 & 39.55 & 38.2 & 41.27 \\
        Deepseek-V3-0324 & 75.22 & 67.27 & 86.59 & 47.71 & 60.97 & 56.00 & 90.16 & 40.45 & 46.03 & 61.90 & 35.65 & 61.58 & 46.86 & 44.58 & 47.17  \\
        DeepSeek-V3.1 & 77.02 & 70.20 & 88.97 & 48.67 & 62.37 & 55.22 & 91.19 & 33.60 & 46.42 & 62.08 & 31.94 & 58.8 & 41.64 & 39.77 & 43.04 \\

        \midrule
        \multicolumn{16}{c}{\textit{The Scale of Large Language Models $<$ 10B}} \\
        \midrule
        
        Qwen3-4B & \underline{69.80} & 60.88 & 81.90 & 38.31 & 34.69 & 44.50 & 79.27 & 28.17 & 46.75 & 51.81 & 26.87 & 54.26 & 36.35 & 35.41 & 38.22 \\
        Qwen2.5-7B & 63.01 & 49.50 & 73.80 & 42.37 & 45.32 & 45.41 & 88.08 & 33.76 & 44.65 & 52.86 & 31.43 & 55.91 & 38.36 & 36.48 & 40.55 \\
        Llama-3.1-8B & 51.24 & 37.74 & 66.62 & 33.32 & 31.27 & 47.10 & 74.61 & 26.88 & 38.60 & 44.52 & 23.07 & 48.15 & 29.32 & 29.13 & 32.42 \\
        Ministral-8B & 49.93 & 42.62 & 70.58 & 36.24 & 30.71 & 37.79 & 82.38 & 28.04 & 46.27 & 46.83 & 25.67 & 50.91 & 32.02 & 31.18 & 34.95 \\
        InternLM3-8B & 58.55 & 51.83 & 76.98 & 38.65 & 25.25 & 39.41 & 66.84 & 44.71 & 43.46 & 48.39 & 24.85 & 50.44 & 35.58 & 34.04 & 36.23 \\
        Qwen3-8B & 66.90 & 58.67 & 82.44 & 46.47 & 48.45 & 44.68 & 89.12 & 27.95 & 45.89 & 55.46 & 33.21 & 58.81 & 40.09 & 38.85 & 42.74 \\
        GLM-4-9B-0414 & 63.27 & 56.03 & 77.67 & 38.03 & 45.29 & 47.01 & 51.30 & 27.51 & 45.52 & 48.55 & 32.20 & 56.90 & 39.73 & 37.40 & 41.57 \\
        RedOne-7B & 63.83 & 72.18 & 88.02 & 65.09 & 63.98 & 51.86 & 70.47 & 74.73 & 48.69 & \underline{66.88} & 38.06 & 62.66 & 46.88 & 44.82 & \textbf{48.11} \\
        \rowcolor{blue!5} \textbf{RedOne 2.0 4B} & \textbf{70.80} & 75.85 & 89.05 & 60.92 & 66.54 & 43.15 & 78.76 & 79.11 & 47.17 & \textbf{67.57} & 38.61 & 62.46 & 45.78 & 43.84 & \underline{47.67} \\
        
        \midrule
        \multicolumn{16}{c}{\textit{10B $<$ The Scale of Large Language Models $<$ 100B}} \\
        \midrule
        
        Phi-4-14B & 63.00 & 57.62 & 79.56 & 46.32 & 53.39 & 44.99 & 89.12 & 29.23 & 44.76 & 55.62 & 31.28 & 57.23 & 37.58 & 36.68 & 40.69 \\
        GPT-OSS-20B & \underline{74.76} & 62.89 & 83.99 & 54.58 & 56.43 & 54.81 & 92.23 & 32.68 & 45.26 & 60.36 & 30.74 & 57.46 & 37.83 & 36.19 & 40.56 \\
        Mistral-Small-24B & 65.63 & 64.88 & 83.89 & 48.77 & 46.51 & 52.09 & 91.19 & 32.10 & 46.01 & 58.18 & 31.29 & 56.72 & 39.28 & 37.32 & 41.15 \\
        Qwen3-30B-A3B & 74.46 & 64.29 & 85.81 & 44.75 & 52.23 & 45.75 & 90.16 & 27.19 & 45.67 & 56.98 & 34.07 & 58.86 & 41.19 & 39.51 & 37.05 \\
        GLM-4-32B-0414 & 74.39 & 63.36 & 85.50 & 47.33 & 53.72 & 50.41 & 80.31 & 33.19 & 46.90 & 57.59 & 36.32 & 61.31 & 42.53 & 40.77 & 45.23 \\
        Qwen2.5-32B & 71.68 & 59.90 & 80.51 & 46.00 & 55.04 & 54.51 & 90.67 & 38.84 & 45.66 & 58.89 & 32.56 & 58.14 & 42.34 & 40.71 & 43.44 \\
        Qwen3-32B & 72.67 & 61.52 & 86.04 & 49.39 & 54.56 & 53.76 & 91.19 & 33.48 & 45.74 & 59.46 & 32.15 & 58.54 & 40.44 & 38.85 & 42.50 \\
        Llama-3.3-70B & 67.64 & 62.94 & 83.28 & 50.76 & 27.38 & 56.09 & 91.19 & 33.58 & 46.41 & 56.45 & 34.00 & 59.18 & 41.25 & 39.56 & 43.50 \\
        RedOne-32B & 73.72 & 81.45 & 90.19 & 67.07 & 59.24 & 51.66 & 81.87 & 70.40 & 50.37 & \underline{69.03} & 40.55 & 64.54 & 48.20 & 46.05 & \textbf{49.84} \\
        \rowcolor{blue!5} 
        \textbf{RedOne 2.0 30B-A3B} & \textbf{75.17} & 77.02 & 89.99 & 63.76 & 62.16 & 54.15 & 81.87 & 74.19 & 49.15 & \textbf{69.04} & 40.22 & 63.88 & 48.06 & 45.95 & \underline{49.54} \\
        
        \bottomrule
        \end{tabular}%
    }
\end{table*}
\subsection{Refinement Learning}
The final stage, Refinement Learning, consolidates prior gains and achieves further performance improvements. This is done by applying RL after the previous SFT-based stage, with the training again centered on SNS data. 

\subsubsection{Further Refinement}
Specifically, we use approximately 400K examples drawn from the SNS and general sources as in the previous stage, with an emphasis on the difficult subsets. We initialize the policy from the prior stage to provide a strong starting base model, and then apply preference-based DAPO~\cite{yu2025dapo} as same as the first stage training process for refinement. In this stage, we also increase the proportion of samples with rationale to 57.18\%, further preserving the model's reasoning ability and benefiting a broad range of downstream tasks. After training, the model's behavior is stabilized and smoothed within the explored solution space, yielding further improvements on both SNS-specific and general tasks. Compared to the previous stage, the RL-based refinement delivers better convergence and more robust domain adaptation.

\section{Experiments}
\subsection{Implementation Details}
During the Exploratory Learning stage, we trained for 500 steps with maximum prompt/response lengths of 10,000/8,192 tokens (18,192 total), plus a 4,096-token overlong buffer with 1.0 penalty factor. We used a prompt batch size of 1,024 with 16 responses per prompt (global batch size 16,384) and mini-batch size covering 256 prompts, yielding 4 gradient updates per rollout.
We adopted DAPO with clipping parameters $\varepsilon_{\mathrm{low}}=0.2$ and $\varepsilon_{\mathrm{high}}=0.28$. Optimization employed AdamW with a constant learning rate of $5\times10^{-6}$, weight decay 0.1, with linear warmup applied for 10 rollout steps. In Targeted Fine-Tuning, we trained for 2 epochs with batch size 64 and maximum sequence length of 16,384 using sequence packing. We optimized cross-entropy loss with AdamW at a learning rate $5\times10^{-6}$, applying a warmup ratio of 0.1 followed by cosine scheduling. The final Refinement Learning stage mirrored the first stage configuration.

\subsection{Experimental Setting}
\subsubsection{Benchmarks}
We perform a comprehensive evaluation of RedOne 2.0 and baselines in both the general and SNS domain capabilities using commonly used benchmarks in the community. Specifically, in general domain, we systematically assess six capabilities, including knowledge reasoning, mathematical reasoning, code generation, machine translation, instruction following, and hallucination detection, as well as \emph{CompassBench}~\cite{2023opencompass}, a comprehensive bench to provide an integrated, multi-dimensional view of model performance. 1) \textbf{Knowledge Reasoning.} We use \emph{MMLU}~\cite{mmlu}, \emph{CMMLU}~\cite{li2023cmmlu}, \emph{C-Eval}~\cite{ceval}, \emph{GPQA-Diamond}~\cite{rein2024gpqa}, \emph{NewsBench}~\cite{li2024newsbench}, \emph{MMLU-Pro}~\cite{mmlupro}, \emph{BBH}~\cite{bbh}, and \emph{GaokaoBench}~\cite{gaokaobench} to probe broad and specialized knowledge, reasoning robustness, difficulty-calibrated multiple choice, and exam-style generalization in both English and Chinese.
2) \textbf{Mathematical Reasoning.} We adopt \emph{GSM8K}~\cite{gsm8k}, \emph{MATH500}~\cite{math}, and the high-stakes \emph{AIME 2025}~\cite{aime25} set to measure multi-step arithmetic and competition-level problem solving.
3) \textbf{Code Generation.} We evaluate program synthesis and correctness with \emph{HumanEval}~\cite{HumanEval}, \emph{MBPP}~\cite{mbpp}, and the temporally refreshed, contamination-aware \emph{LiveCodeBench}~\cite{jain2024livecodebench}, reporting pass@k and execution-based metrics.
4) \textbf{Machine Translation.} We benchmark multilingual translation with the WMT tasks (i.e. WMT-22~\cite{WMT22}, WMT-23~\cite{wmt23} and WMT-24~\cite{wmt24}) and \emph{FLORES}~\cite{flores}, covering diverse language pairs and domains.
5) \textbf{Instruction Following.} We employ \emph{IFEval}~\cite{ifeval}, which provides automatically verifiable constraints to quantify compliance under explicit instructions.
6) \textbf{Hallucination Detection.} We use \emph{HaluEval}~\cite{halueval} to assess the tendency to produce unverifiable or fabricated content across question answering, dialogue, and summarization settings.

In the SNS domain, we validate models on benchmarks built from real SNS scenarios, covering five aspects: post comprehension, information retrieval, sentiment and intent analysis, personalized recommendation, and translation. We use SNS-Bench~\cite{sns-bench}, a large-scale bench with 6,658 questions spanning eight tasks from a social platform with over 300M users, which includes the following tasks: 1) \textbf{Note-Taxonomy (Taxon.)} for content categorization; 2) \textbf{Note-Hashtag (Hash.)} to select suitable tags; 3) \textbf{Note-QueryCorr (QCorr)} to align user queries with note content and topic; 4) \textbf{Note-MRC (MRC)} for simple and complex reading comprehension over long notes; 5) \textbf{Note-NER (NER)} for entity extraction; 6) \textbf{Note-Gender (Gender)} to assess gender-sensitive appeal; 7) \textbf{Note-CHLW (CHLW)} to highlight salient words in comment threads; and 8) \textbf{Note-QueryGen (QGen)} to produce effective search queries. For translation, we adopt SNS-TransBench~\cite{redtrans-bench}, a curated set of 2,858 English–Chinese cases from posts, comments, and multimedia captions that emphasizes phenomena central to SNS translation, including humor localization, emoji semantics, and meme adaptation. It tests whether models can preserve pragmatics, style, and culture-bound references in short, high-context text typical of social platforms.

\subsubsection{Baselines}
We conduct comparison experiments with various proprietary models, including GPT4o-1120~\cite{gpt4o}, Gemini-2.0-Flash~\cite{team2023gemini}, Claude-3.7-Sonnet~\cite{claude3.7}, Doubao-1.5-Pro-32k~\cite{doubao_1_5_pro}, Qwen-Max~\cite{qwen2.5}, and GLM-4-Plus~\cite{glm2024chatglm}, open-source models, such as Qwen series~\cite{qwen2.5,yang2025qwen3}, Llama series~\cite{grattafiori2024llama}, Ministral~\cite{ministral}, Mistral-Small-24B~\cite{mistralsmall2025}, InternLM3-8B~\cite{internlm}, Phi-4-14B~\cite{abdin2024phi}, dots.llm1~\cite{dotsllm1}, gpt-oss series~\cite{gptoss}, GLM series~\cite{glm4.5} and DeepSeek series~\cite{deepseekv3}, as well as SNS domain specific models RedOne~\cite{zhao2025redone}. 


\subsection{Main Results}
As shown in Table~\ref{tab:main_table}, we conduct a comparison with various models across General-Bench, SNS-Bench, and SNS-TransBench, covering a broad spectrum of capabilities from general reasoning to SNS-domain understanding and multilingual transfer. 

Across all benchmarks, RedOne 2.0 averagely achieves strong and balanced results, surpassing both open- and closed-source baselines of comparable scale. Specifically, RedOne 2.0 4B attains the highest average score on General-Bench with 70.8, exceeding even larger open models such as Qwen3-8B and GLM-4-9B, and achieving comparable or superior results to some proprietary LLMs or LLMs with more than 100B parameters. This demonstrates that the proposed three-stage post-training pipeline effectively enhances both general and domain-specific capabilities even at smaller scales. On SNS-Bench, which evaluates domain-specific understanding and reasoning across eight tasks, RedOne 2.0 still leads within its scale group. The 4B variant achieves an average score of 67.57, outperforming all sub-10B baselines and exceeding the previous RedOne-7B model by 0.69, despite having fewer parameters. Similarly, the 30B-A3B version achieves 69.04, even matching or surpassing much larger models such as GPT-4o and GLM-4.5. These results validate that RedOne 2.0 not only inherits strong generalization from the base models but also substantially improves SNS-domain competence through progressive alignment. On SNS-TransBench, which measures cross-lingual understanding and translation quality between Chinese and English, RedOne 2.0 maintains competitive results across BLEU and chrF++ metrics. Both the 4B and 30B-A3B variants achieve the top-2 overall averages with 47.67 and 49.54, respectively, outperforming all similarly scaled models. The consistent performance across both translation directions indicates that RedOne 2.0's alignment pipeline preserves linguistic versatility while improving domain adaptation.

We further observe clear scalability in the RedOne 2.0 family. As model size increases from 4B to 30B, both general and SNS-specific metrics steadily improve, with the average gain on General-Becnh, SNS-Bench and SNS-TransBench exceeding 4.37, 1.47 and 1.87, respectively. This confirms that the proposed three-stage pipeline scales effectively and provides stable improvements without overfitting to the SNS domain. Compared with the RedOne models, RedOne 2.0 shows promising gains across majority evaluation suites. For instance, RedOne 2.0 4B improves by 6.97 on General-Bench and by 0.69 on SNS-Bench relative to RedOne-7B. Additionally, for SNS-TransBench, although RedOne 2.0 series are smaller than RedOne, they still achieve comparable performance. And comparing with RedOne, RedOne 2.0 obtains higher improvement from base model. Overall, these results confirm the effectiveness of RedOne 2.0 in achieving efficient, scalable and stable alignment.


\begin{table}[t]
    \centering
    \caption{Generalization of our training pipeline over different base models. We report the average performance for General-Bench, SNS-Bench and SNS-TransBench.}
    \label{tab:model_scales}
    \resizebox{\linewidth}{!}{%
    \begin{tabular}{l|ccc}
        \toprule
        \textbf{Models} & \textbf{General-Bench} & \textbf{SNS-Bench} & \textbf{SNS-TransBench} \\
        \midrule
        Qwen3-4B & 69.80 & 51.81 & 38.22 \\
        RedOne 2.0 4B & 70.80 & 67.57 & 47.67 \\
        \midrule
        Qwen3-8B & 66.90 & 55.46 & 42.74 \\
        RedOne 2.0 8B & 69.27 & 65.82 & 46.72\\
        \midrule
        Qwen3-30B-A3B & 74.46 & 56.98 & 37.05 \\
        RedOne 2.0 30B-A3B & 75.17 & 69.04 & 49.54 \\
        \midrule
        Qwen3-32B & 72.67 & 59.46 & 42.50 \\
        RedOne 2.0 32B & 73.17 & 69.76 & 49.11  \\
        \bottomrule
        \end{tabular}%
    }
\end{table}
\subsection{More Analysis}
\subsubsection{Generalization Across Different Base-Model Scales.}
Table~\ref{tab:model_scales} demonstrates that our three-stage training pipeline generalizes effectively across base models of different scales. Consistent improvements are observed on all benchmarks, confirming the robustness of our approach in transferring alignment benefits from smaller to larger models. Moreover, scaling up the base model further amplifies the overall performance of RedOne~2.0, indicating that larger capacities allow more effective utilization of the staged optimization signals. Notably, the 4B and 30B-A3B variants exhibit superior gains comparing with similar scale models, particularly on SNS-related benchmarks. We think this is because that they are both based on instruction-tuned backbones, and stronger instruction-following capabilities can better absorb and express the multi-stage alignment signals, leading to stronger generalization and domain adaptation.

\begin{table}[t]
    \centering
    \caption{The impact of different training stage on Qwen3-4B's performance.}
    \label{tab:incremental_perform}    
    \resizebox{\linewidth}{!}{
    \begin{tabular}{ccc|ccc}
        \toprule
        \multirow{2}{*}{\makecell[c]{Exploratory\\Learning}} & \multirow{2}{*}{\makecell[c]{Targeted\\Fine-Tuning}} & \multirow{2}{*}{\makecell[c]{Refinement\\Learning}} & \multirow{2}{*}{\makecell[c]{General\\-Bench}} & \multirow{2}{*}{\makecell[c]{SNS\\-Bench}} & \multirow{2}{*}{\makecell[c]{SNS-Trans\\Bench}} \\
         & & & & & \\
        \midrule
         & & & 69.80 & 51.81 & 38.22 \\
         & \textbf{$\checkmark$} & & 63.65 & 61.10 & 46.00 \\
         & \textbf{$\checkmark$} & \textbf{$\checkmark$} & 69.80 & 63.03 & 45.95\\
        \textbf{$\checkmark$} & & & 71.25 & 62.27& 43.35\\
        \textbf{$\checkmark$} & \textbf{$\checkmark$} & & 70.04 & 65.67 & 47.72\\
        \textbf{$\checkmark$} & \textbf{$\checkmark$} & \textbf{$\checkmark$} & 70.80 & 67.57 & 47.67 \\
        \bottomrule
    \end{tabular}%
    }
\end{table}

\begin{table}[t]
    \centering
    \caption{Comparison with task specific fine-tuning on Qwen3-4B and RedOne2.0 4B.}
    \label{tab:cmp_tsft}
    \resizebox{.95\linewidth}{!}{%
    \begin{tabular}{c|ccc}
        \toprule
        \textbf{Models} & \textbf{Hash.} & \textbf{QCorr} & \textbf{MRC}   \\
        \midrule
        Qwen3-4B & 81.90 & 38.31 & 34.69 \\
        Qwen3-4B (Fine-tuned) & 90.12 & 60.11 & 57.54 \\
        RedOne 2.0 4B & 89.05 & 60.92 & 66.54 \\
        \midrule
        \textbf{Models} & \textbf{CHLW} & \textbf{QGen} & \textbf{SNS-Trans} \\
        \midrule
        Qwen3-4B & 28.17 & 46.75 & 38.22 \\
        Qwen3-4B (Fine-tuned) & 67.24 & 49.24 & 44.25 \\
        RedOne 2.0 4B & 79.11 & 47.17 & 47.67 \\
        \bottomrule
        \end{tabular}%
    }
\end{table}
\subsubsection{Incremental Performance Over three-Stage Training.}
As shown in Table~\ref{tab:incremental_perform}), we evaluate the incremental impact of each stage in our three-phase training framework. The RL-based Exploratory Learning stage establishes a strong foundation, improving performance to 71.25\% on General-Bench, 62.27\% on SNS-Bench, and 43.35\% on SNS-TransBench, highlighting its effectiveness in consistently enhancing the overall capability of the base model. The SFT-based Targeted Fine-Tuning stage then addresses weaknesses in the SNS domain exhibited in previous stage, raising scores to 65.67\% on SNS-Bench and 47.72\% on SNS-TransBench, with only a slight drop 1.21\% on General-Bench. Finally, the RL-based Refinement Learning stage balances performance across tasks, increasing the average from 61.14\% to 62.01\% and resulting in final scores of 70.80\% on General-Bench, 67.57\% on SNS-Bench, and 47.67\% on SNS-TransBench.

\subsubsection{Superiority to the Naive SFT Followed RL Baseline.}
Then, considering the most notable shift in RedOne 2.0 lies in its departure from the traditional SFT-centric domain-specific post-training paradigm to RL, we conduct experiments to compare it with naive SFT followed RL baseline. This baseline typically began with SFT for domain adaptation, followed by RL to align the model with human preferences or downstream objectives, as shown in the second and third rows of Table~\ref{tab:incremental_perform}. While SFT can effectively boost performance in SNS domains, it often causes a ``seesaw'' effect, significantly reducing general capability from 69.80 to 63.65. Although subsequent RL attempts to mitigate this issue, the overall improvements across the three benchmarks remain limited. In contrast, RedOne 2.0 refine the process: starting with RL to establish domain priors, followed by SFT for targeted enhancement, and concluding with RL for final optimization. This paradigm effectively avoids the trade-off between general and domain-specific performance and surpasses the naive baseline by 1.00 on General-Bench, 4.54 on SNS-Bench, and 1.72 on SNS-TransBench.

\subsubsection{Comparison with Task Specific Fine-tuning.}
We also compares RedOne2.0 framework, designed for unified optimization across all tasks, against the conventional task specific fine-tuning approach. As detailed in Table \ref{tab:cmp_tsft}, this method also yields strong performance on its target objective. For instance, a Qwen3-4B model fine-tuned specifically for QGen achieves 49.24, and another fine-tuned for Hash. reaches 90.12. However, RedOne2.0 4B, trained concurrently on a mixture of all tasks, delivers robust and highly competitive results across the entire spectrum of benchmarks. Especially, it outperforms the task specific fine-tuned Qwen3-4B models on MRC with 9.00, and on CHLW with 11.87. It also maintains strong performance on QCorr at 60.92 and on SNS-Trans at 47.67. These results substantiate that a unified training framework can effectively capture and leverage beneficial inter-task relationships, enabling a single model to achieve comprehensive and better capability.


\begin{table}[t]
    \centering
    \caption{RedOne2.0's online application on personalized re-creation of posts' title.}
    \label{tab:online}
    \resizebox{\linewidth}{!}{%
    \begin{tabular}{cc|c}
        \toprule
        \multicolumn{2}{c|}{\textbf{Metrics}} & {\textbf{Relative Change}} \\
        \midrule
        Business Value & Advertiser Value (AdvV) $\uparrow$ & +0.43\% \\
        \midrule
        \multirow{4}{*}{Content Quality} & Vague Titles Ratio $\downarrow$ & -11.9\% \\
         & Practical Titles Ratio $\uparrow$ & +7.1\% \\
         & Authentic Titles Ratio $\uparrow$ & +12.9\% \\
         & Interactive Titles Ratio $\uparrow$ & +25.8\% \\
        \bottomrule
        \end{tabular}%
    }
\end{table}
\subsection{Online Application}
We deployed RedOne2.0 on a large scale social networking platform with more than 3 million users to recommend personalized re-created post titles in real time. Each pre-published title is routed to the service, which performs semantic analysis and produces an enhanced title that preserves the original intent while optimizing for engagement. The suggestion is exhibited to the creator or, in selected traffic buckets, used directly to measure performance against the original. Evaluation covered business impact and content quality. The primary business indicator is Advertiser Value (AdvV), which reflects the value delivered to advertisers through audience quality and engagement. Content quality is measured through human review across four dimensions: vagueness, practicality, authenticity, and interactivity.

As shown in Table~\ref{tab:online}, the online test was conducted over several weeks and millions of posts, and showed consistent gains. Advertiser Value increased by 0.43\%, a statistically significant improvement at platform scale. Human evaluation reported an 11.9\% reduction in vague titles and increases of 7.1\% in practical titles, 12.9\% in authentic titles, and 25.8\% in interactive titles. The strong rise in interactive titles indicates that the model learns linguistic patterns that encourage responses such as comments and shares. These results also suggest a clear link between quality improvements and business outcomes. Moreover, practical and authentic titles are likely to increase user trust and dwell time, while interactive titles stimulate community activity. Deploying RedOne2.0 therefore improves user experience and yields measurable advertiser value, which demonstrates its effectiveness for real world content optimization.

\begin{table}[t]
    \centering
    \caption{Good cases for personalized re-creation of post titles.}
    \label{tab:case_study_good}
    \resizebox{\linewidth}{!}{%
    \begin{tabular}{c|c}
        \toprule
        \textbf{Title Source} & \textbf{Content} \\
        
        \midrule
        \multicolumn{2}{c}{\textit{Case 1}} \\
        \midrule
        
        \multirow{2}{*}{Original} & \multirow{2}{*}{\makecell[c]{Plum rain season, a great helper\\for dehumidification and mold prevention.}} \\
        {} & {} \\
        \cmidrule{1-2}
        \multirow{2}{*}{Base Model} & \multirow{2}{*}{\makecell[c]{Dehumidification essential for the plum rain\\season,a fresh choice for a dry and comfortable life.}} \\
        {} & {} \\
        \cmidrule{1-2}
        \multirow{2}{*}{RedOne 2.0} & \multirow{2}{*}{\makecell[c]{Say goodbye to ``steamy'' homes!\\Rescue your plum rain season.}} \\
        {} & {} \\
        
        \midrule
        \multicolumn{2}{c}{\textit{Case 2}} \\
        \midrule
        
        \multirow{2}{*}{Original} & \multirow{2}{*}{\makecell[c]{Beijing wedding photo recommendations:\\17 lawn wedding photo outdoor spots.}} \\
        {} & {} \\
        \cmidrule{1-2}
        \multirow{2}{*}{Base Model} & \multirow{2}{*}{\makecell[c]{Dreamy lawn wedding photos,\\capturing the most beautiful moments.}} \\
        {} & {} \\
        \cmidrule{1-2}
        \multirow{2}{*}{RedOne 2.0} & \multirow{2}{*}{\makecell[c]{Escape the studio! 17 stunning\\lawns capture cinematic-level wedding photos.}} \\
        {} & {} \\
        \bottomrule
        \end{tabular}%
    }
\end{table}

\begin{table}[t]
    \centering
    \caption{Bad case for personalized re-creation of post titles.}
    \label{tab:case_study_bad}
    \resizebox{\linewidth}{!}{%
    \begin{tabular}{c|c}
        \toprule
        \textbf{Title Source} & \textbf{Content} \\
        \midrule
        \multirow{2}{*}{Original} & \multirow{2}{*}{\makecell[c]{Don't buy the wrong transportation card for\\Osaka and Kyoto! A lesson learned the hard way!}} \\
        {} & {} \\
        \cmidrule{1-2}
        \multirow{2}{*}{Base Model} & \multirow{2}{*}{\makecell[c]{A guide to Japanese transportation\\cards—stop making these mistakes!}} \\
        {} & {} \\
        \cmidrule{1-2}
        \multirow{2}{*}{RedOne 2.0} & \multirow{2}{*}{\makecell[c]{Avoid these pitfalls for your Kansai trip,\\check out the guide now.}} \\
        {} & {} \\
        \bottomrule
        \end{tabular}%
    }
\end{table}
\subsection{Case Study}
To qualitatively assess RedOne 2.0, we compare its outputs with a baseline on personalized re-created titles. As shown in Table~\ref{tab:case_study_good}, RedOne 2.0 consistently produces more evocative and engaging phrasing. For the dehumidification example, the baseline remains serviceable yet generic, whereas RedOne 2.0 introduces a vivid word of a steamy home and adds a clear imperative that heightens emotional resonance. A similar pattern holds for the example of the wedding photography. The baseline provides a broad description, while RedOne 2.0 frames the content as an exclusive discovery that promises cinematic results, which is likely to raise curiosity and strengthen click intent. We also exhibited a bad case in scenarios that require strict preservation of critical facts, as illustrated in Table~\ref{tab:case_study_bad}. The original title centers on the risk of choosing the wrong transportation card for Osaka and Kyoto and stresses a hard learned lesson. The baseline remains close to this focus. RedOne 2.0 generates a more interactive sentence but generalizes the topic to Kansai travel and omits the key reference to the transportation card, which weakens informational precision and urgency. These cases indicate that RedOne 2.0 excels at optimizing for engagement and stylistic appeal yet can sometimes over optimize at the expense of essential details. Future work should reinforce faithfulness constraints while preserving expressiveness.

\section{Conclusion}
In this paper, we present RedOne 2.0, an SNS-specific LLM post-training framework tailored for SNS, where tasks are highly heterogeneous, dynamic, and culturally diverse. Unlike traditional SFT-centric approaches that risk catastrophic forgetting and unstable trade-offs between in-domain and out-of-domain performance, RedOne 2.0 adopts a progressive, RL-prioritized three-stage pipeline: Exploratory Learning to establish initial domain alignment and surface weaknesses, Targeted Fine-Tuning to selectively repair deficiencies while retaining general competence, and Refinement Learning to consolidate improvements via RL with SNS-centric rewards. Supported by a large, task-diverse dataset spanning more than 75 SNS tasks and high quality general corpus, this paradigm demonstrates strong data efficiency, stable adaptation, and robust generalization even at compact model scales. Overall, RedOne 2.0 provides a competitive, cost-effective, and scalable baseline for LLM deployment in SNS, advancing capability without sacrificing robustness, safety, or general usability.



\bibliographystyle{ACM-Reference-Format}
\bibliography{reference}

@String{Computing = "Computing" }

@String{Chelsea = "Chelsea" }

@String{Springer = "Springer-Verlag" }

@ArtifactSoftware{R,
    title = {R: A Language and Environment for Statistical Computing},
    author = {{R Core Team}},
    organization = {R Foundation for Statistical Computing},
    address = {Vienna, Austria},
    year = {2019},
    url = {https://www.R-project.org/},
}

@article{jiang2023socialllm,
  title={Social-llm: Modeling user behavior at scale using language models and social network data},
  author={Jiang, Julie and Ferrara, Emilio},
  journal={arXiv preprint arXiv:2401.00893},
  year={2023}
}

@article{zeng2024llmforsocialnetworks,
  title={Large language models for social networks: Applications, challenges, and solutions},
  author={Zeng, Jingying and Huang, Richard and Malik, Waleed and Yin, Langxuan and Babic, Bojan and Shacham, Danny and Yan, Xiao and Yang, Jaewon and He, Qi},
  journal={arXiv preprint arXiv:2401.02575},
  year={2024}
}

@article{zhao2025redone,
  title={RedOne: Revealing Domain-specific LLM Post-Training in Social Networking Services},
  author={Zhao, Fei and Lu, Chonggang and Wang, Yue and Xie, Zheyong and Liu, Ziyan and Qian, Haofu and Huang, JianZhao and Shi, Fangcheng and Meng, Zijie and Guo, Hongcheng and others},
  journal={arXiv preprint arXiv:2507.10605},
  year={2025}
}

@article{bakshy2015exposure,
  title={Exposure to ideologically diverse news and opinion on Facebook},
  author={Bakshy, Eytan and Messing, Solomon and Adamic, Lada A},
  journal={Science},
  volume={348},
  number={6239},
  pages={1130--1132},
  year={2015},
  publisher={American Association for the Advancement of Science}
}

@article{vosoughi2018spread,
  title={The spread of true and false news online},
  author={Vosoughi, Soroush and Roy, Deb and Aral, Sinan},
  journal={science},
  volume={359},
  number={6380},
  pages={1146--1151},
  year={2018},
  publisher={American Association for the Advancement of Science}
}

@article{altay2025following,
  title={Following news on social media boosts knowledge, belief accuracy and trust},
  author={Altay, Sacha and Hoes, Emma and Wojcieszak, Magdalena},
  journal={Nature Human Behaviour},
  pages={1--10},
  year={2025},
  publisher={Nature Publishing Group}
}

@article{zhang2023sentiment,
  title={Sentiment analysis in the era of large language models: A reality check},
  author={Zhang, Wenxuan and Deng, Yue and Liu, Bing and Pan, Sinno Jialin and Bing, Lidong},
  journal={arXiv preprint arXiv:2305.15005},
  year={2023}
}

@inproceedings{wang2023reducing,
  title={Reducing spurious correlations in aspect-based sentiment analysis with explanation from large language models},
  author={Wang, Qianlong and Ding, Keyang and Liang, Bin and Yang, Min and Xu, Ruifeng},
  booktitle={Findings of the Association for Computational Linguistics: EMNLP 2023},
  pages={2930--2941},
  year={2023}
}

@inproceedings{lubos2024llm,
  title={LLM-generated explanations for recommender systems},
  author={Lubos, Sebastian and Tran, Thi Ngoc Trang and Felfernig, Alexander and Polat Erdeniz, Seda and Le, Viet-Man},
  booktitle={Adjunct Proceedings of the 32nd ACM Conference on User Modeling, Adaptation and Personalization},
  pages={276--285},
  year={2024}
}

@article{lyu2023llm,
  title={Llm-rec: Personalized recommendation via prompting large language models},
  author={Lyu, Hanjia and Jiang, Song and Zeng, Hanqing and Xia, Yinglong and Wang, Qifan and Zhang, Si and Chen, Ren and Leung, Christopher and Tang, Jiajie and Luo, Jiebo},
  journal={arXiv preprint arXiv:2307.15780},
  year={2023}
}

@inproceedings{kumar2024watch,
  title={Watch your language: Investigating content moderation with large language models},
  author={Kumar, Deepak and AbuHashem, Yousef Anees and Durumeric, Zakir},
  booktitle={Proceedings of the International AAAI Conference on Web and Social Media},
  volume={18},
  pages={865--878},
  year={2024}
}

@inproceedings{kolla2024llm,
  title={Llm-mod: Can large language models assist content moderation?},
  author={Kolla, Mahi and Salunkhe, Siddharth and Chandrasekharan, Eshwar and Saha, Koustuv},
  booktitle={Extended Abstracts of the CHI Conference on Human Factors in Computing Systems},
  pages={1--8},
  year={2024}
}

@article{feng2024does,
  title={What does the bot say? opportunities and risks of large language models in social media bot detection},
  author={Feng, Shangbin and Wan, Herun and Wang, Ningnan and Tan, Zhaoxuan and Luo, Minnan and Tsvetkov, Yulia},
  journal={arXiv preprint arXiv:2402.00371},
  year={2024}
}

@inproceedings{qiao2025botsim,
  title={BotSim: LLM-Powered Malicious Social Botnet Simulation},
  author={Qiao, Boyu and Li, Kun and Zhou, Wei and Li, Shilong and Lu, Qianqian and Hu, Songlin},
  booktitle={Proceedings of the AAAI Conference on Artificial Intelligence},
  volume={39},
  number={13},
  pages={14377--14385},
  year={2025}
}

@article{shah2024stackeval,
  title={Stackeval: Benchmarking llms in coding assistance},
  author={Shah, Nidhish and Genc, Zulkuf and Araci, Dogu},
  journal={Advances in Neural Information Processing Systems},
  volume={37},
  pages={36976--36994},
  year={2024}
}

@inproceedings{kahl2024llms,
  title={LLMs Cannot (Yet) Match the Specificity and Simplicity of Online Communities in Long Form Question Answering},
  author={Kahl, Kris-Fillip and Buz, Tolga and Biswas, Russa and De Melo, Gerard},
  booktitle={Findings of the Association for Computational Linguistics: EMNLP 2024},
  pages={2028--2053},
  year={2024}
}

@article{wu2023bloomberggpt,
  title={Bloomberggpt: A large language model for finance},
  author={Wu, Shijie and Irsoy, Ozan and Lu, Steven and Dabravolski, Vadim and Dredze, Mark and Gehrmann, Sebastian and Kambadur, Prabhanjan and Rosenberg, David and Mann, Gideon},
  journal={arXiv preprint arXiv:2303.17564},
  year={2023}
}

@article{bhatia2024fintral,
  title={Fintral: A family of gpt-4 level multimodal financial large language models},
  author={Bhatia, Gagan and Nagoudi, El Moatez Billah and Cavusoglu, Hasan and Abdul-Mageed, Muhammad},
  journal={arXiv preprint arXiv:2402.10986},
  year={2024}
}

@article{yang2023investlm,
  title={Investlm: A large language model for investment using financial domain instruction tuning},
  author={Yang, Yi and Tang, Yixuan and Tam, Kar Yan},
  journal={arXiv preprint arXiv:2309.13064},
  year={2023}
}

@article{gruca2008incentive,
  title={Incentive and accuracy issues in movie prediction markets},
  author={Gruca, Thomas S and Berg, Joyce E and Cipriano, Michael},
  journal={The Journal of Prediction Markets},
  volume={2},
  number={1},
  pages={29--43},
  year={2008}
}

@article{guha2023legalbench,
  title={Legalbench: A collaboratively built benchmark for measuring legal reasoning in large language models},
  author={Guha, Neel and Nyarko, Julian and Ho, Daniel and R{\'e}, Christopher and Chilton, Adam and Chohlas-Wood, Alex and Peters, Austin and Waldon, Brandon and Rockmore, Daniel and Zambrano, Diego and others},
  journal={Advances in neural information processing systems},
  volume={36},
  pages={44123--44279},
  year={2023}
}

@article{katz2024gpt,
  title={Gpt-4 passes the bar exam},
  author={Katz, Daniel Martin and Bommarito, Michael James and Gao, Shang and Arredondo, Pablo},
  journal={Philosophical Transactions of the Royal Society A},
  volume={382},
  number={2270},
  pages={20230254},
  year={2024},
  publisher={The Royal Society}
}

@article{dahl2024large,
  title={Large legal fictions: Profiling legal hallucinations in large language models},
  author={Dahl, Matthew and Magesh, Varun and Suzgun, Mirac and Ho, Daniel E},
  journal={Journal of Legal Analysis},
  volume={16},
  number={1},
  pages={64--93},
  year={2024},
  publisher={Oxford University Press UK}
}

@article{colombo2024saullm,
  title={Saullm-7b: A pioneering large language model for law},
  author={Colombo, Pierre and Pires, Telmo Pessoa and Boudiaf, Malik and Culver, Dominic and Melo, Rui and Corro, Caio and Martins, Andre FT and Esposito, Fabrizio and Raposo, Vera L{\'u}cia and Morgado, Sofia and others},
  journal={arXiv preprint arXiv:2403.03883},
  year={2024}
}

@article{singhal2025toward,
  title={Toward expert-level medical question answering with large language models},
  author={Singhal, Karan and Tu, Tao and Gottweis, Juraj and Sayres, Rory and Wulczyn, Ellery and Amin, Mohamed and Hou, Le and Clark, Kevin and Pfohl, Stephen R and Cole-Lewis, Heather and others},
  journal={Nature Medicine},
  volume={31},
  number={3},
  pages={943--950},
  year={2025},
  publisher={Nature Publishing Group US New York}
}

@article{xu2025lingshu,
  title={Lingshu: A Generalist Foundation Model for Unified Multimodal Medical Understanding and Reasoning},
  author={Xu, Weiwen and Chan, Hou Pong and Li, Long and Aljunied, Mahani and Yuan, Ruifeng and Wang, Jianyu and Xiao, Chenghao and Chen, Guizhen and Liu, Chaoqun and Li, Zhaodonghui and others},
  journal={arXiv preprint arXiv:2506.07044},
  year={2025}
}

@article{liu2025dentvlm,
  title={DentVLM: A Multimodal Vision-Language Model for Comprehensive Dental Diagnosis and Enhanced Clinical Practice},
  author={Meng, Zijie and Hao, Jin and Dai, Xiwei and Feng, Yang and Liu, Jiaxiang and Feng, Bin and Wu, Huikai and Gai, Xiaotang and Zhu, Hengchuan and Hu, Tianxiang and others},
  journal={arXiv preprint arXiv:2509.23344},
  year={2025}
}

@article{jiang2025omniv,
  title={Omniv-med: Scaling medical vision-language model for universal visual understanding},
  author={Jiang, Songtao and Wang, Yuan and Song, Sibo and Zhang, Yan and Meng, Zijie and Lei, Bohan and Wu, Jian and Sun, Jimeng and Liu, Zuozhu},
  journal={arXiv preprint arXiv:2504.14692},
  year={2025}
}

@article{zhang2024sciglm,
  title={Sciglm: Training scientific language models with self-reflective instruction annotation and tuning},
  author={Zhang, Dan and Hu, Ziniu and Zhoubian, Sining and Du, Zhengxiao and Yang, Kaiyu and Wang, Zihan and Yue, Yisong and Dong, Yuxiao and Tang, Jie},
  journal={arXiv preprint arXiv:2401.07950},
  year={2024}
}

@article{azerbayev2023llemma,
  title={Llemma: An open language model for mathematics},
  author={Azerbayev, Zhangir and Schoelkopf, Hailey and Paster, Keiran and Santos, Marco Dos and McAleer, Stephen and Jiang, Albert Q and Deng, Jia and Biderman, Stella and Welleck, Sean},
  journal={arXiv preprint arXiv:2310.10631},
  year={2023}
}

@article{taylor2022galactica,
  title={Galactica: A large language model for science},
  author={Taylor, Ross and Kardas, Marcin and Cucurull, Guillem and Scialom, Thomas and Hartshorn, Anthony and Saravia, Elvis and Poulton, Andrew and Kerkez, Viktor and Stojnic, Robert},
  journal={arXiv preprint arXiv:2211.09085},
  year={2022}
}

@article{bai2025intern,
  title={Intern-s1: A scientific multimodal foundation model},
  author={Bai, Lei and Cai, Zhongrui and Cao, Maosong and Cao, Weihan and Chen, Chiyu and Chen, Haojiong and Chen, Kai and Chen, Pengcheng and Chen, Ying and Chen, Yongkang and others},
  journal={arXiv preprint arXiv:2508.15763},
  year={2025}
}

@article{yu2025dapo,
  title={Dapo: An open-source llm reinforcement learning system at scale},
  author={Yu, Qiying and Zhang, Zheng and Zhu, Ruofei and Yuan, Yufeng and Zuo, Xiaochen and Yue, Yu and Dai, Weinan and Fan, Tiantian and Liu, Gaohong and Liu, Lingjun and others},
  journal={arXiv preprint arXiv:2503.14476},
  year={2025}
}

@article{deepseekr1,
  title={Deepseek-r1: Incentivizing reasoning capability in llms via reinforcement learning},
  author={Guo, Daya and Yang, Dejian and Zhang, Haowei and Song, Junxiao and Zhang, Ruoyu and Xu, Runxin and Zhu, Qihao and Ma, Shirong and Wang, Peiyi and Bi, Xiao and others},
  journal={arXiv preprint arXiv:2501.12948},
  year={2025}
}

@misc{2023opencompass,
    title={OpenCompass: A Universal Evaluation Platform for Foundation Models},
    author={OpenCompass Contributors},
    howpublished = {\url{https://github.com/open-compass/opencompass}},
    year={2023}
}

@article{mmlu,
  title={Measuring massive multitask language understanding},
  author={Hendrycks, Dan and Burns, Collin and Basart, Steven and Zou, Andy and Mazeika, Mantas and Song, Dawn and Steinhardt, Jacob},
  journal={arXiv preprint arXiv:2009.03300},
  year={2020}
}

@article{li2023cmmlu,
  title={Cmmlu: Measuring massive multitask language understanding in chinese},
  author={Li, Haonan and Zhang, Yixuan and Koto, Fajri and Yang, Yifei and Zhao, Hai and Gong, Yeyun and Duan, Nan and Baldwin, Timothy},
  journal={arXiv preprint arXiv:2306.09212},
  year={2023}
}

@article{ceval,
  title={C-eval: A multi-level multi-discipline chinese evaluation suite for foundation models},
  author={Huang, Yuzhen and Bai, Yuzhuo and Zhu, Zhihao and Zhang, Junlei and Zhang, Jinghan and Su, Tangjun and Liu, Junteng and Lv, Chuancheng and Zhang, Yikai and Fu, Yao and others},
  journal={Advances in Neural Information Processing Systems},
  volume={36},
  pages={62991--63010},
  year={2023}
}

@inproceedings{rein2024gpqa,
  title={Gpqa: A graduate-level google-proof q\&a benchmark},
  author={Rein, David and Hou, Betty Li and Stickland, Asa Cooper and Petty, Jackson and Pang, Richard Yuanzhe and Dirani, Julien and Michael, Julian and Bowman, Samuel R},
  booktitle={First Conference on Language Modeling},
  year={2024}
}

@article{li2024newsbench,
  title={NewsBench: a systematic evaluation framework for assessing editorial capabilities of large language models in chinese journalism},
  author={Li, Miao and Chen, Ming-Bin and Tang, Bo and Hou, Shengbin and Wang, Pengyu and Deng, Haiying and Li, Zhiyu and Xiong, Feiyu and Mao, Keming and Cheng, Peng and others},
  journal={arXiv preprint arXiv:2403.00862},
  year={2024}
}

@article{mmlupro,
  title={Mmlu-pro: A more robust and challenging multi-task language understanding benchmark},
  author={Wang, Yubo and Ma, Xueguang and Zhang, Ge and Ni, Yuansheng and Chandra, Abhranil and Guo, Shiguang and Ren, Weiming and Arulraj, Aaran and He, Xuan and Jiang, Ziyan and others},
  journal={Advances in Neural Information Processing Systems},
  volume={37},
  pages={95266--95290},
  year={2024}
}

@article{bbh,
  title={Challenging big-bench tasks and whether chain-of-thought can solve them},
  author={Suzgun, Mirac and Scales, Nathan and Sch{\"a}rli, Nathanael and Gehrmann, Sebastian and Tay, Yi and Chung, Hyung Won and Chowdhery, Aakanksha and Le, Quoc V and Chi, Ed H and Zhou, Denny and others},
  journal={arXiv preprint arXiv:2210.09261},
  year={2022}
}

@article{gaokaobench,
  title={Evaluating the performance of large language models on gaokao benchmark},
  author={Zhang, Xiaotian and Li, Chunyang and Zong, Yi and Ying, Zhengyu and He, Liang and Qiu, Xipeng},
  journal={arXiv preprint arXiv:2305.12474},
  year={2023}
}

@article{aime25,
  author       = {{MAA}},
  title        = {American Invitational Mathematics Examination - AIME},
  journal      = {American Invitational Mathematics Examination - AIME 2025},
  year         = {2025},
  url          = {https://maa.org/math-competitions/american-invitational-mathematics-examination-aime}
}

@article{gsm8k,
  title={Training verifiers to solve math word problems},
  author={Cobbe, Karl and Kosaraju, Vineet and Bavarian, Mohammad and Chen, Mark and Jun, Heewoo and Kaiser, Lukasz and Plappert, Matthias and Tworek, Jerry and Hilton, Jacob and Nakano, Reiichiro and others},
  journal={arXiv preprint arXiv:2110.14168},
  year={2021}
}

@article{math,
  title={Measuring mathematical problem solving with the math dataset},
  author={Hendrycks, Dan and Burns, Collin and Kadavath, Saurav and Arora, Akul and Basart, Steven and Tang, Eric and Song, Dawn and Steinhardt, Jacob},
  journal={arXiv preprint arXiv:2103.03874},
  year={2021}
}

@article{HumanEval,
  title={Evaluating large language models trained on code},
  author={Chen, Mark and Tworek, Jerry and Jun, Heewoo and Yuan, Qiming and Pinto, Henrique Ponde De Oliveira and Kaplan, Jared and Edwards, Harri and Burda, Yuri and Joseph, Nicholas and Brockman, Greg and others},
  journal={arXiv preprint arXiv:2107.03374},
  year={2021}
}

@article{mbpp,
  title={Program synthesis with large language models},
  author={Austin, Jacob and Odena, Augustus and Nye, Maxwell and Bosma, Maarten and Michalewski, Henryk and Dohan, David and Jiang, Ellen and Cai, Carrie and Terry, Michael and Le, Quoc and others},
  journal={arXiv preprint arXiv:2108.07732},
  year={2021}
}

@article{jain2024livecodebench,
  title={Livecodebench: Holistic and contamination free evaluation of large language models for code},
  author={Jain, Naman and Han, King and Gu, Alex and Li, Wen-Ding and Yan, Fanjia and Zhang, Tianjun and Wang, Sida and Solar-Lezama, Armando and Sen, Koushik and Stoica, Ion},
  journal={arXiv preprint arXiv:2403.07974},
  year={2024}
}

@inproceedings{WMT22,
  title={Findings of the 2022 conference on machine translation (WMT22)},
  author={Kocmi, Tom and Bawden, Rachel and Bojar, Ond{\v{r}}ej and Dvorkovich, Anton and Federmann, Christian and Fishel, Mark and Gowda, Thamme and Graham, Yvette and Grundkiewicz, Roman and Haddow, Barry and others},
  booktitle={Proceedings of the Seventh Conference on Machine Translation (WMT)},
  pages={1--45},
  year={2022}
}

@inproceedings{wmt23,
  title={Results of WMT23 metrics shared task: Metrics might be guilty but references are not innocent},
  author={Freitag, Markus and Mathur, Nitika and Lo, Chi-kiu and Avramidis, Eleftherios and Rei, Ricardo and Thompson, Brian and Kocmi, Tom and Blain, Frederic and Deutsch, Daniel and Stewart, Craig and others},
  booktitle={Proceedings of the Eighth Conference on Machine Translation},
  pages={578--628},
  year={2023}
}

@inproceedings{wmt24,
  title={Findings of the WMT24 general machine translation shared task: The LLM era is here but MT is not solved yet},
  author={Kocmi, Tom and Avramidis, Eleftherios and Bawden, Rachel and Bojar, Ond{\v{r}}ej and Dvorkovich, Anton and Federmann, Christian and Fishel, Mark and Freitag, Markus and Gowda, Thamme and Grundkiewicz, Roman and others},
  booktitle={Proceedings of the Ninth Conference on Machine Translation},
  pages={1--46},
  year={2024}
}

@article{flores,
  title={The flores-101 evaluation benchmark for low-resource and multilingual machine translation},
  author={Goyal, Naman and Gao, Cynthia and Chaudhary, Vishrav and Chen, Peng-Jen and Wenzek, Guillaume and Ju, Da and Krishnan, Sanjana and Ranzato, Marc’Aurelio and Guzm{\'a}n, Francisco and Fan, Angela},
  journal={Transactions of the Association for Computational Linguistics},
  volume={10},
  pages={522--538},
  year={2022},
  publisher={MIT Press One Broadway, 12th Floor, Cambridge, Massachusetts 02142, USA~…}
}

@article{ifeval,
  title={Instruction-following evaluation for large language models},
  author={Zhou, Jeffrey and Lu, Tianjian and Mishra, Swaroop and Brahma, Siddhartha and Basu, Sujoy and Luan, Yi and Zhou, Denny and Hou, Le},
  journal={arXiv preprint arXiv:2311.07911},
  year={2023}
}

@article{halueval,
  title={Halueval: A large-scale hallucination evaluation benchmark for large language models},
  author={Li, Junyi and Cheng, Xiaoxue and Zhao, Wayne Xin and Nie, Jian-Yun and Wen, Ji-Rong},
  journal={arXiv preprint arXiv:2305.11747},
  year={2023}
}

@inproceedings{sns-bench,
  title={SNS-Bench: Defining, Building, and Assessing Capabilities of Large Language Models in Social Networking Services},
  author={Guo, Hongcheng and Cao, Shaosheng and Wang, Boyang and Li, Lei and Chen, Liang and Lyu, Xinze and Xu, Zhe and Hu, Yao and Li, Zhoujun and others},
  booktitle={Forty-second International Conference on Machine Learning}
}

@article{redtrans-bench,
  title={Redefining Machine Translation on Social Network Services with Large Language Models},
  author={Guo, Hongcheng and Zhao, Fei and Cao, Shaosheng and Lyu, Xinze and Liu, Ziyan and Wang, Yue and Wang, Boyang and Li, Zhoujun and Lu, Chonggang and Xu, Zhe and others},
  journal={arXiv preprint arXiv:2504.07901},
  year={2025}
}

@article{gpt4o,
  title={Gpt-4o system card},
  author={Hurst, Aaron and Lerer, Adam and Goucher, Adam P and Perelman, Adam and Ramesh, Aditya and Clark, Aidan and Ostrow, AJ and Welihinda, Akila and Hayes, Alan and Radford, Alec and others},
  journal={arXiv preprint arXiv:2410.21276},
  year={2024}
}

@article{gptoss,
  title={gpt-oss-120b \& gpt-oss-20b model card},
  author={Agarwal, Sandhini and Ahmad, Lama and Ai, Jason and Altman, Sam and Applebaum, Andy and Arbus, Edwin and Arora, Rahul K and Bai, Yu and Baker, Bowen and Bao, Haiming and others},
  journal={arXiv preprint arXiv:2508.10925},
  year={2025}
}

@article{glm4.5,
  title={Glm-4.5: Agentic, reasoning, and coding (arc) foundation models},
  author={Zeng, Aohan and Lv, Xin and Zheng, Qinkai and Hou, Zhenyu and Chen, Bin and Xie, Chengxing and Wang, Cunxiang and Yin, Da and Zeng, Hao and Zhang, Jiajie and others},
  journal={arXiv preprint arXiv:2508.06471},
  year={2025}
}

@article{deepseekv3,
  title={Deepseek-v3 technical report},
  author={Liu, Aixin and Feng, Bei and Xue, Bing and Wang, Bingxuan and Wu, Bochao and Lu, Chengda and Zhao, Chenggang and Deng, Chengqi and Zhang, Chenyu and Ruan, Chong and others},
  journal={arXiv preprint arXiv:2412.19437},
  year={2024}
}

@article{qwen2.5,
  title={Qwen2. 5 Technical Report},
  author={Yang, An and Yang, Baosong and Zhang, Beichen and Hui, Binyuan and Zheng, Bo and Yu, Bowen and Li, Chengyuan and Liu, Dayiheng and Huang, Fei and Wei, Haoran and others},
  journal={CoRR},
  year={2024}
}

@article{yang2025qwen3,
  title={Qwen3 technical report},
  author={Yang, An and Li, Anfeng and Yang, Baosong and Zhang, Beichen and Hui, Binyuan and Zheng, Bo and Yu, Bowen and Gao, Chang and Huang, Chengen and Lv, Chenxu and others},
  journal={arXiv preprint arXiv:2505.09388},
  year={2025}
}

@article{team2023gemini,
  title={Gemini: a family of highly capable multimodal models},
  author={Team, Gemini and Anil, Rohan and Borgeaud, Sebastian and Alayrac, Jean-Baptiste and Yu, Jiahui and Soricut, Radu and Schalkwyk, Johan and Dai, Andrew M and Hauth, Anja and Millican, Katie and others},
  journal={arXiv preprint arXiv:2312.11805},
  year={2023}
}

@article{dotsllm1,
  title={dots. llm1 Technical Report},
  author={Huo, Bi and Tu, Bin and Qin, Cheng and Zheng, Da and Zhang, Debing and Zhang, Dongjie and Li, En and Guo, Fu and Yao, Jian and Lou, Jie and others},
  journal={arXiv preprint arXiv:2506.05767},
  year={2025}
}

@misc{claude3.7,
  title={3.7 sonnet and claude code},
  author={Anthropic, Claude},
  year={2025}
}

@misc{doubao_1_5_pro,
  author       = {Doubao-Team},
  title        = {Doubao-1.5-pro: Model release},
  year         = {2025},
  month        = {January},
  howpublished = {\url{https://team.doubao.com/en/special/doubao_1_5_pro}},
}

@article{abdin2024phi,
  title={Phi-4 technical report},
  author={Abdin, Marah and Aneja, Jyoti and Behl, Harkirat and Bubeck, S{\'e}bastien and Eldan, Ronen and Gunasekar, Suriya and Harrison, Michael and Hewett, Russell J and Javaheripi, Mojan and Kauffmann, Piero and others},
  journal={arXiv preprint arXiv:2412.08905},
  year={2024}
}

@article{grattafiori2024llama,
  title={The llama 3 herd of models},
  author={Grattafiori, Aaron and Dubey, Abhimanyu and Jauhri, Abhinav and Pandey, Abhinav and Kadian, Abhishek and Al-Dahle, Ahmad and Letman, Aiesha and Mathur, Akhil and Schelten, Alan and Vaughan, Alex and others},
  journal={arXiv preprint arXiv:2407.21783},
  year={2024}
}

@misc{ministral,
  author       = {Mistral-AI},
  title        = {Un Ministral, des Ministraux},
  howpublished = {\url{https://mistral.ai/news/ministraux}},
  note         = {Accessed: 2024-10-16},
  year         = {2024},
  month        = {Oct},
}

@misc{mistralsmall2025,
  author       = {Mistral-AI},
  title        = {Mistral Small 3.1},
  howpublished = {\url{https://mistral.ai/news/mistral-small-3-1}},
  note         = {Accessed: 2025-03-17},
  year         = {2025},
  month        = {Mar},
}

@misc{internlm,
      title={InternLM2 Technical Report},
      author={Zheng Cai and Maosong Cao and Haojiong Chen and Kai Chen and Keyu Chen and Xin Chen and Xun Chen and Zehui Chen and Zhi Chen and Pei Chu and Xiaoyi Dong and Haodong Duan and Qi Fan and Zhaoye Fei and Yang Gao and Jiaye Ge and Chenya Gu and Yuzhe Gu and Tao Gui and Aijia Guo and Qipeng Guo and Conghui He and Yingfan Hu and Ting Huang and Tao Jiang and Penglong Jiao and Zhenjiang Jin and Zhikai Lei and Jiaxing Li and Jingwen Li and Linyang Li and Shuaibin Li and Wei Li and Yining Li and Hongwei Liu and Jiangning Liu and Jiawei Hong and Kaiwen Liu and Kuikun Liu and Xiaoran Liu and Chengqi Lv and Haijun Lv and Kai Lv and Li Ma and Runyuan Ma and Zerun Ma and Wenchang Ning and Linke Ouyang and Jiantao Qiu and Yuan Qu and Fukai Shang and Yunfan Shao and Demin Song and Zifan Song and Zhihao Sui and Peng Sun and Yu Sun and Huanze Tang and Bin Wang and Guoteng Wang and Jiaqi Wang and Jiayu Wang and Rui Wang and Yudong Wang and Ziyi Wang and Xingjian Wei and Qizhen Weng and Fan Wu and Yingtong Xiong and Chao Xu and Ruiliang Xu and Hang Yan and Yirong Yan and Xiaogui Yang and Haochen Ye and Huaiyuan Ying and Jia Yu and Jing Yu and Yuhang Zang and Chuyu Zhang and Li Zhang and Pan Zhang and Peng Zhang and Ruijie Zhang and Shuo Zhang and Songyang Zhang and Wenjian Zhang and Wenwei Zhang and Xingcheng Zhang and Xinyue Zhang and Hui Zhao and Qian Zhao and Xiaomeng Zhao and Fengzhe Zhou and Zaida Zhou and Jingming Zhuo and Yicheng Zou and Xipeng Qiu and Yu Qiao and Dahua Lin},
      year={2024},
      eprint={2403.17297},
      archivePrefix={arXiv},
      primaryClass={cs.CL}
}

@article{glm2024chatglm,
  title={Chatglm: A family of large language models from glm-130b to glm-4 all tools},
  author={GLM, Team and Zeng, Aohan and Xu, Bin and Wang, Bowen and Zhang, Chenhui and Yin, Da and Zhang, Dan and Rojas, Diego and Feng, Guanyu and Zhao, Hanlin and others},
  journal={arXiv preprint arXiv:2406.12793},
  year={2024}
}

@article{ouyang2022training,
  title={Training language models to follow instructions with human feedback},
  author={Ouyang, Long and Wu, Jeffrey and Jiang, Xu and Almeida, Diogo and Wainwright, Carroll and Mishkin, Pamela and Zhang, Chong and Agarwal, Sandhini and Slama, Katarina and Ray, Alex and others},
  journal={Advances in neural information processing systems},
  volume={35},
  pages={27730--27744},
  year={2022}
}

@article{yuan2023rrhf,
  title={Rrhf: Rank responses to align language models with human feedback without tears},
  author={Yuan, Zheng and Yuan, Hongyi and Tan, Chuanqi and Wang, Wei and Huang, Songfang and Huang, Fei},
  journal={arXiv preprint arXiv:2304.05302},
  year={2023}
}

@article{rafailov2023direct,
  title={Direct preference optimization: Your language model is secretly a reward model},
  author={Rafailov, Rafael and Sharma, Archit and Mitchell, Eric and Manning, Christopher D and Ermon, Stefano and Finn, Chelsea},
  journal={Advances in neural information processing systems},
  volume={36},
  pages={53728--53741},
  year={2023}
}

@article{shao2024deepseekmath,
  title={Deepseekmath: Pushing the limits of mathematical reasoning in open language models},
  author={Shao, Zhihong and Wang, Peiyi and Zhu, Qihao and Xu, Runxin and Song, Junxiao and Bi, Xiao and Zhang, Haowei and Zhang, Mingchuan and Li, YK and Wu, Yang and others},
  journal={arXiv preprint arXiv:2402.03300},
  year={2024}
}

@article{xia2013socially,
  title={Socially aware networking: A survey},
  author={Xia, Feng and Liu, Li and Li, Jie and Ma, Jianhua and Vasilakos, Athanasios V},
  journal={IEEE Systems Journal},
  volume={9},
  number={3},
  pages={904--921},
  year={2013},
  publisher={IEEE}
}

@article{huang2024mitigating,
  title={Mitigating catastrophic forgetting in large language models with self-synthesized rehearsal},
  author={Huang, Jianheng and Cui, Leyang and Wang, Ante and Yang, Chengyi and Liao, Xinting and Song, Linfeng and Yao, Junfeng and Su, Jinsong},
  journal={arXiv preprint arXiv:2403.01244},
  year={2024}
}

@article{kumar2022fine,
  title={Fine-tuning can distort pretrained features and underperform out-of-distribution},
  author={Kumar, Ananya and Raghunathan, Aditi and Jones, Robbie and Ma, Tengyu and Liang, Percy},
  journal={arXiv preprint arXiv:2202.10054},
  year={2022}
}

@article{kotha2023understanding,
  title={Understanding catastrophic forgetting in language models via implicit inference},
  author={Kotha, Suhas and Springer, Jacob Mitchell and Raghunathan, Aditi},
  journal={arXiv preprint arXiv:2309.10105},
  year={2023}
}

@inproceedings{ramasesh2021effect,
  title={Effect of scale on catastrophic forgetting in neural networks},
  author={Ramasesh, Vinay Venkatesh and Lewkowycz, Aitor and Dyer, Ethan},
  booktitle={International conference on learning representations},
  year={2021}
}

@article{jin2025rl,
  title={RL Fine-Tuning Heals OOD Forgetting in SFT},
  author={Jin, Hangzhan and Luan, Sitao and Lyu, Sicheng and Rabusseau, Guillaume and Rabbany, Reihaneh and Precup, Doina and Hamdaqa, Mohammad},
  journal={arXiv preprint arXiv:2509.12235},
  year={2025}
}

@article{li2017learning,
  title={Learning without forgetting},
  author={Li, Zhizhong and Hoiem, Derek},
  journal={IEEE transactions on pattern analysis and machine intelligence},
  volume={40},
  number={12},
  pages={2935--2947},
  year={2017},
  publisher={IEEE}
}

\appendix

\end{document}